\documentclass[]{spie}  %>>> use for US letter paper
\usepackage{amsmath,amsfonts,amssymb}
\usepackage{graphicx}
\usepackage{tabularx}
\usepackage[colorlinks=true, allcolors=blue]{hyperref}
\usepackage{subcaption}
\usepackage{adjustbox}
\usepackage{booktabs}

\newcolumntype{C}{>{\centering\arraybackslash}X}

\title{Temporally Consistent Mitral Annulus Measurements from Sparse Annotations in Echocardiographic Videos}

\author[a,b,c,*]{Gino E. Jansen}
\author[d]{Mark J. Schuuring}
\author[e]{Berto J. Bouma}
\author[a,b,c,f]{Ivana Išgum}
\affil[a]{Department of Biomedical Engineering \& Physics, Amsterdam University Medical Center, Netherlands}
\affil[b]{Informatics Institute, University of Amsterdam, Netherlands}
\affil[c]{Amsterdam Cardiovascular Sciences, Netherlands}
\affil[d]{Department of Biomedical Signals and Systems, University of Twente, Netherlands}
\affil[e]{Department of Cardiology, Amsterdam University Medical Center, Netherlands}
\affil[f]{Department of Radiology \& Nuclear Medicine, Amsterdam University Medical Center, Netherlands}

\authorinfo{Further author information: Send correspondence to Gino E. Jansen. (E-mail: g.e.jansen@amsterdamumc.nl)}

\begin{document} 
\maketitle

\begin{abstract}
This work presents a novel approach to achieving temporally consistent mitral annulus landmark localization in echocardiography videos using sparse annotations. Our method introduces a self-supervised loss term that enforces temporal consistency between neighboring frames, which smooths the position of landmarks and enhances measurement accuracy over time. Additionally, we incorporate realistic field-of-view augmentations to improve the recognition of missing anatomical landmarks. We evaluate our approach on both a public and private dataset, and demonstrate significant improvements in Mitral Annular Plane Systolic Excursion (MAPSE) calculations and overall landmark tracking stability. The method achieves a mean absolute MAPSE error of 1.81 ± 0.14 mm, an annulus size error of 2.46 ± 0.31 mm, and a landmark localization error of 2.48 ± 0.07 mm. Finally, it achieves a 0.99 ROC-AUC for recognition of missing landmarks.
\end{abstract}


\section{INTRODUCTION}
\label{sec:intro}  % \label{} allows reference to this section

% Quantitative measurement of ventricle function and valve dimensions is an integral part of a cardiac ultrasound exam. For example, the mitral annulus dimensions -- derived from two cardiac landmarks -- are crucial for measuring mitral annular dilation. Additionally, the same landmarks enable the calculation of the mitral annular plane excursion (MAPSE) for myocardial function assessment, proposed as a surrogate measure for left-ventricular ejection fraction when image quality is suboptimal.\cite{matos2012mitral} Automatic detection of these landmarks may reduce observer variability and decrease the burden on healthcare professionals analyzing the heart. 

The mitral valve plays a crucial role in cardiac function: It regulates blood flow between the left atrium and left ventricle. Accurate assessment of the mitral annulus -- the fibrous ring supporting the mitral valve leaflets -- is essential for diagnosing and managing various cardiac conditions. Changes in the diameter of the mitral annulus can indicate mitral regurgitation -- unwanted backflow of blood from the left ventricle to the atrium. This condition often arises as a complication of coronary heart disease (CHD), a leading cause of death responsible for about one-third of all deaths in individuals over 35 years old \cite{Sanchis2016}. Another key metric derived from the mitral annulus is the Mitral Annular Plane Systolic Excursion (MAPSE), which measures the longitudinal displacement of the mitral annulus during systole. MAPSE serves as a surrogate marker for left ventricular ejection fraction (LVEF), providing insights into left ventricular systolic function \cite{matos2012mitral}, and it is of vital importance for diagnosing coronary heart disease. Reduced MAPSE values have been associated with increased mortality risk, independent of left ventricular ejection fraction\cite{Romano2017}.

Transthoracic echocardiography is the first-line imaging modality for evaluating the mitral annulus and calculating MAPSE. This requires localization of the two annulus landmarks at end-systole and end-diastole -- time points that could be derived automatically if the anatomical landmarks are tracked accurately and consistently across sequential frames. Ensuring temporal consistency in these measurements is therefore crucial for providing meaningful clinical assessments. Recent advancements in deep learning have enabled supervised algorithms to perform landmark detection in medical images, learning from annotated examples\cite{Schwendicke2021}. However, manual frame-by-frame annotation of these landmarks is labor-intensive and prone to inconsistency across consecutive frames, highlighting the need for automated deep learning methods that ensure both accuracy and temporal consistency, without requiring a dense annotation strategy. Additionally, due to the limited field of view provided by ultrasound imaging, the target anatomy may often disappear because of unintended movements of the patient or operator during acquisition. An automatic landmark localization method needs to be able to account for this by recognizing when the target moves out of view.

Several landmark detection methods have previously been developed for landmark detection in echocardiography, in particular for measuring left-ventricle wall thickness \cite{Gilbert2019, Howard2021, Duffy2022}. This task requires localization of landmarks along the boundaries of the septal and posterior wall in parasternal long-axis (PLAX) view. In these works, the authors employed fully convolutional networks -- such as U-Net, DeeplabV3, or HRNet -- to predict high-resolution heatmaps, whose centroids represent the predicted landmark location. These methods reach high performance on distance measures, but lack evaluations on temporal consistency, as their main focus lies with replicating the (sparse) annotations made by human observers. 
Several studies have addressed the challenge of temporally sparse annotations for deep-learning landmark detection in echocardiography videos \cite{Lin2021, Goco2022}, but only for detecting landmarks on the inner wall of the left ventricle in PLAX-view. Furthermore, Smistad et al. \cite{smistad2022tracking} proposed a framework for landmark-tracking-based MAPSE calculation. This method firstly used a network that selects end-diastolic (ED) and end-systolic (ES) frames from a full video, then used another neural network to detect the annulus landmarks in ED, and propagated these landmarks over time with a third network that predicted the optical flow field between every two consecutive frames. Through landmark propagation, spatiotemporal relations were accounted for, but this also caused error accumulation.

To incorporate spatiotemporal relations into automated landmark localization, we present a novel approach that incorporates consistency with neighboring frames, while requiring sparsely available landmark annotations only. Additionally, we propose training augmentations that improve the model's ability to recognize missing (out-of-view) landmarks. We evaluate the method on two echocardiography datasets: a public dataset with left-heart segmentations of the apical 2- and 4-chamber view from which we extract mitral annulus landmarks; and a private dataset, which additionally includes apical 3-chamber views.

\section{Data}
The datasets used in this study include the publicly available CAMUS dataset \cite{Leclerc2019}, and a private dataset of echocardiography videos retrospectively collected from Amsterdam University Medical Center (AUMC). The annotations in both datasets include sparsely annotated left-ventricle contours, from which we extract the endpoints as target mitral annulus landmarks. 

The CAMUS dataset contains a 1,000 videos from 500 patients, with 49\% of patients having a pathological ejection fraction below $45\%$. Each patient study contains two videos showing the systolic phase of the A2C and A4C views, respectively, with each view containing a reference segmentation of the left ventricle for the end-diastolic and end-systolic frame. 

The AUMC dataset contains 293 videos from 104 patients with symptoms of angina and/or dyspnea, which include 100 apical 2-chamber (A2C), 92 apical 3-chamber (A3C), and 101 apical 4-chamber (A4C) views. For each video, several end-diastolic, end-systolic, and random intermediate frames were annotated, resulting in a total of 1,541 annotated frames. 

\begin{figure}[t!]
    \centering
    \includegraphics[width=\linewidth, trim={0 -6cm 0 0},clip]{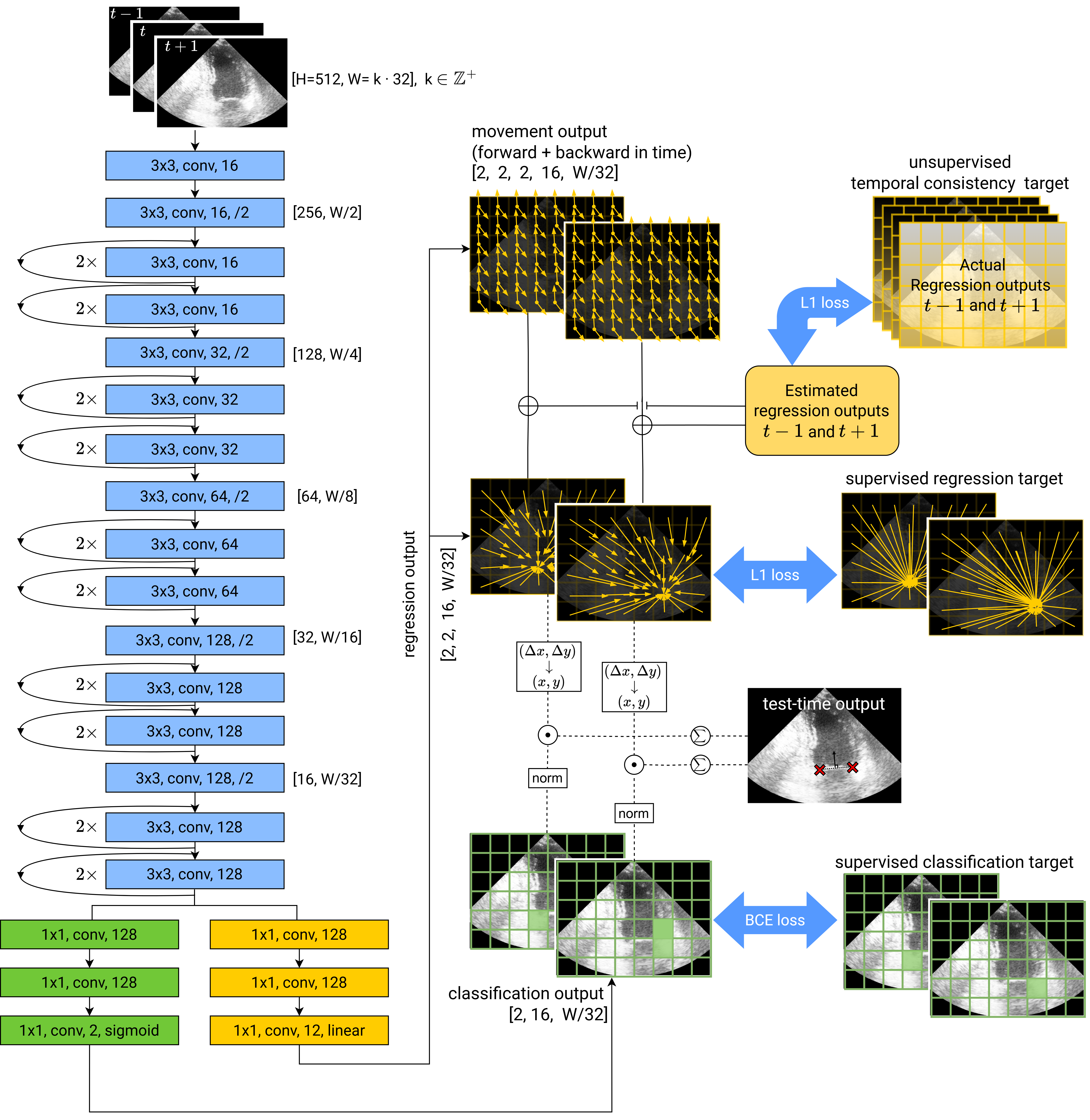}
    %[trim={left bottom right top},clip]
    \caption{Architecture of the proposed method. A ResNet-like fully convolutional neural network takes three consecutive frames as input, stacked in the channel dimension, and outputs classification and regression maps for mitral annulus landmarks. The blue convolutional blocks include GroupNorm, while other blocks omit normalization. ReLU is used as the activation function unless stated otherwise. Temporal consistency is enforced through an unsupervised loss that compares predicted displacements across neighboring frames, while supervised losses guide the regression and classification outputs. During test time, landmarks are computed as the weighted mean of regressed locations from each patch, using the respective probabilities from the classification map.}
    \label{fig:method}
\end{figure}

\section{Method}
\subsection{Landmark detection}
Inspired by the landmark detection method proposed by Noothout et al.\cite{Noothout2020}, we employ a network that applies both classification and regression on a spatially down-sampled feature representation of the input image, as illustrated in Figure \ref{fig:method}. To enable landmark detection in videos, we added another loss term to the training that measures the consistency between neighboring frames. Additionally, we apply realistic cropping and rotation augmentations to improve robustness to missing landmarks.
% The general framework for landmark localization follows the methodology proposed by Noothout et al.\cite{Noothout2020}, that performs both regression and classification on patch-based features.

To localize a landmark, a 2D fully convolutional neural network (CNN) creates a patch-based representation of an input image. Then, a classification head classifies for each patch if the target landmark is located within it, and a regression head infers the (x- and y-) distance to the target from the center-point coordinates of each patch. Finally, landmark coordinates are obtained by calculating the average of regressed locations, weighted by corresponding classification probabilities. To train the network, we optimize the binary cross-entropy loss between the probability map from the classification output and a target map of the same resolution, containing zeros except for the patch containing the landmark, which is set to 1. Additionally, we optimize the L1-distance loss between the regression output's distance map and a target distance map of the same resolution, which represents the log-distance of each patch's center point to the target landmark. 

To achieve temporally consistent video landmark detection, we add a loss term that measures the consistency between neighboring frames. To learn this consistency, the CNN receives three consecutive frames as input, stacked in the channel dimension, denoted as $I_t$, where $\{ I_t \}_{t=1}^{T} \subseteq \mathbb{R}^{3 \times H \times W}$. From this small sequence of images, the CNN infers displacement vectors from the center point of each image patch to the predicted landmark for the current time point $t$, denoted as $d_t \in \{x,y\}$. Additionally, it predicts the positional displacement vectors to neighboring frames at time points ${t-1}$ and ${t+1}$, which we call the forward and backward displacement, denoted as $\Delta d^{-}$ and $\Delta d^{+}$, respectively. In parallel, the CNN performs identical operations on neighboring inputs $I_{t-1}$ and $I_{t+1}$. To compute the consistency loss between time point $t$ and its neighbors, we determine the locations at $t\pm 1$, both directly from inputs $I_{t\pm 1}$, and derived from input $I_{t}$. Then, we compute the temporal consistency loss as the error between those inferred locations: 
\begin{equation}
    \mathcal{L}_{temp} (t) = \sum_{d \in \{x, y\}} | d_{t} + \Delta d_{t}^{+} - d_{t+1} | + | d_{t} + \Delta d_{t}^{-} - d_{t-1} |, 
\end{equation}
where $|\cdot|$ is used to denote the absolute value. We average this loss first for all image patches and then all videos, and we scale it by $\beta(=0.5)$ to weigh the consistency loss before adding it to the other loss terms, which yields:

\begin{equation}
    \mathcal{L}_{total} = \left\langle \mathcal{L}_{C} (t) + \mathcal{L}_{R} (t) \right\rangle_{t \in T_A} + \beta \cdot \left\langle \mathcal{L}_{temp} (t) \right\rangle_{t \in T},
\end{equation}
where \({L}_{C}\) and \({L}_{R}\) are the classification and regression losses, respectively. \( \left\langle \cdot \right\rangle_{t \in T} \) denotes the mean over time points \( t \) in the set \( T \), which are all the frames in a video, whereas \( T_A \) represents the set of annotated frames.

\begin{figure}[t]
    \centering
    \includegraphics[width=0.7\textwidth]{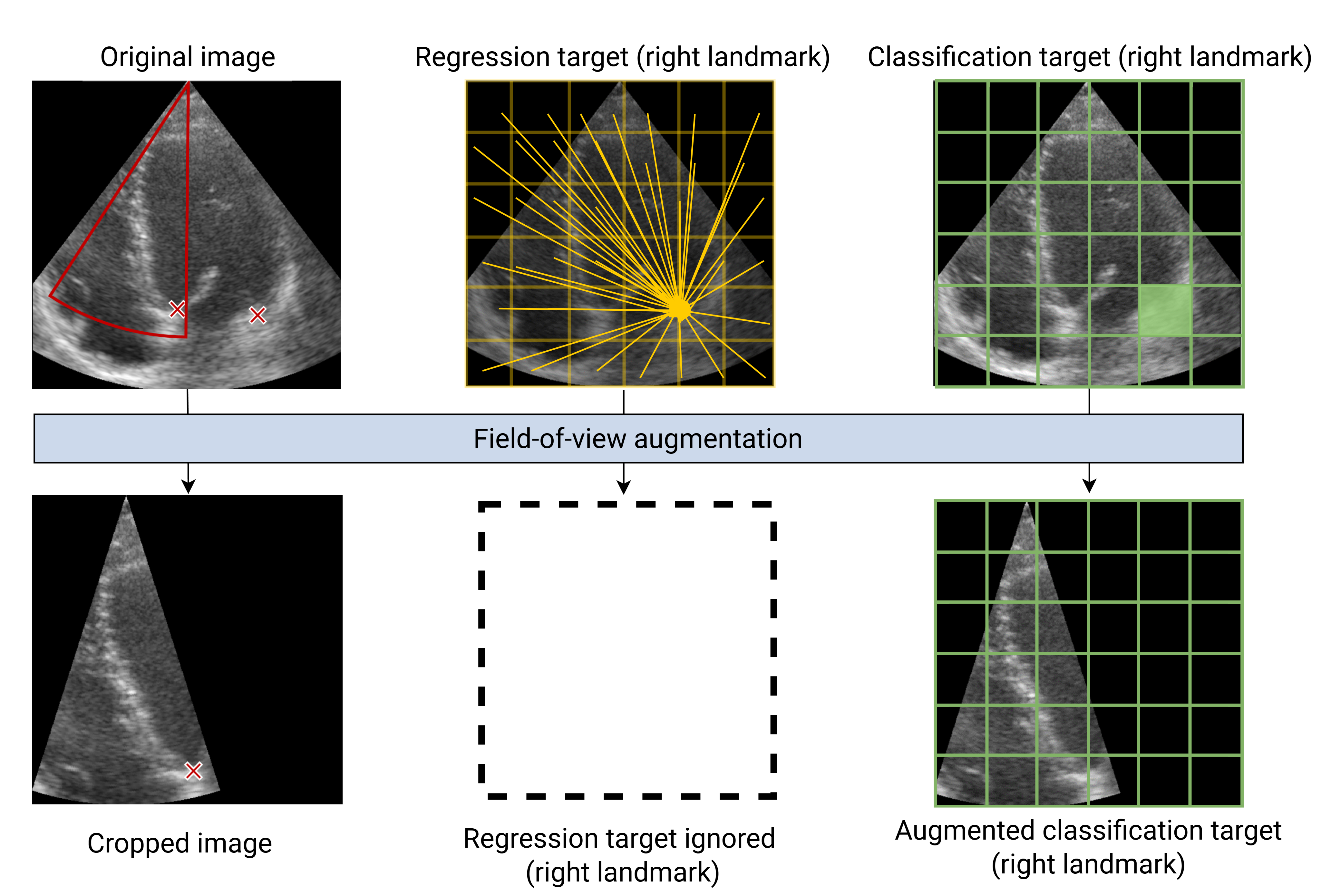} % Replace with your first image file
    \caption{Training image and associated targets for the right annulus landmark undergoing field-of-view augmentation. The original image is cropped using a sector-shaped cropping window, which excludes the right landmark due to random cropping. Consequently, the regression target is ignored and excluded from back-propagation, while the classification target is updated to an array of all zeros to reflect the absence of the landmark in the cropped region.}
    \label{fig:fov_augment}
\end{figure}

Finally, to achieve more robustness to missing landmarks, we apply realistic cropping and rotation augmentations. The model receives randomly augmented videos during training, which may cause a landmark to be moved out of view. This prompts the model to learn to recognize missing anatomical landmarks. The augmentation scheme consists of a zoom-in and a rotation, and is conducted in a manner that preserves the sector-shaped field of view, as illustrated in Figure \ref{fig:fov_augment}. The landmark targets are transformed along with the image: if the landmark falls outside of the sector-shaped field of view, its corresponding target classification map is nullified, whereas the regression output is excluded from back-propagation. During test time, a landmark prediction is withheld if the maximum probability for a landmark in the image fails to exceed a certain threshold. This threshold is set based on the maximum accuracy that the model achieves on another (validation) set with test-time augmentations.

\subsection{Architecture}
To extract spatial features for the landmark localization task, we employed a Resnet-like architecture without a global pooling operation, and we used GroupNorm instead of BatchNorm to enable training with small batches. Among its convolutional layers, there were 5 downsampling layers with stride 2, which reduced the spatial input dimensions from 512$\times$512 to 16$\times$16 pixels in the output. Finally, the regression and classification heads each contained three 1$\times$1 convolutional layers. Further details are illustrated in Figure \ref{fig:method}.

\subsection{Evaluation}
We evaluate performance of the proposed method using the mean absolute errors (MAE) of landmark localization, MAPSE calculation, and mitral annulus size (distance between annulus landmarks). MAPSE is computed as the systolic displacement (ED to ES) of the annular plane along the time-averaged normal of the annular plane \cite{smistad2022tracking}. In the AUMC dataset, videos are excluded from MAPSE computation if they do not include consecutive ED and ES annotations.

As a measure of smoothness, we compute the mean absolute jerk of each landmark over time (mm/frame$^3$), which is the third-order time derivative of its position, indicating sudden changes in acceleration. As the cardiac motion is characterized by smooth transitions in acceleration, jerk mostly reflects measurement errors caused by tracking inconsistencies. 

We compare the proposed approach against the baseline method developed by Noothout et al. \cite{Noothout2020}, specifically focusing on their global-only implementation, which forms the foundation for this work.

\section{Experiments \& Results}

\subsection{Training details}
The CAMUS and AUMC sets were combined, and jointly trained and evaluated on. The CAMUS data was split into a training set of 400 patients (800 videos), a validation set of 50 patients (100 videos), and a test set with the remaining 50 patients (100 videos). For training and testing, we sampled the full half-cycle video with two annotated frames. The AUMC data was randomly split into a training and test set on the patient level, resulting in a split of 235 videos (83 patients) in the training set and 58 videos (21 patients) in the test set. For training, we sampled each annotated frame along with a clip of 30 consecutive frames with the annotated frame in the middle, while for testing we sampled the full video. 

We trained the method using the Adam optimizer with default settings and a learning rate of $0.0001$ and run 300k iterations. We sampled batches of 4 videos, and computed the localization loss on the annotated frames, while computing the temporal consistency loss on all sampled frames in a video. To enable more meaningful comparisons, statistically, we trained the networks in each experiment 5 times using different random seeds. During inference, we did not utilize the movement outputs, and computed the landmark locations using the classification and regression outputs only (see Figure \ref{fig:method}). 
% Prior to utilization, videos are resized in to 512 pixels in height with preserved aspect ratio. This is followed by a data augmentation (in training), after which the image is cropped (or extended) to a width of 512 pixels. Additionally, video pixels are normalized from [0, 255] to [0, 1].

\begin{table}[t!]
\centering
\small
\caption{Metrics comparing the proposed temporally consistent method with the baseline approach\cite{Noothout2020}. Errors are measured in millimeters (mm). Statistics are reported as mean $\pm$ standard deviation across different random seeds. Significant improvements in performance are highlighted in bold font ($p<0.05$, paired t-test).}
\begin{tabularx}{\textwidth}{@{}XCCCC@{}}
\toprule
 & \textbf{MAPSE (MAE)} & \textbf{Annulus Size (MAE)} & \textbf{Landmark Location (MAE)} & \textbf{Mean Jerk}\\ \midrule
  % & \textbf{(MAE)} & \textbf{} & \textbf{(MAE)} & \\ 
\textbf{Proposed} & \textbf{1.81 ± 0.14} & 2.46 ± 0.31 & \textbf{2.48 ± 0.07} & \textbf{0.98 ± 0.01} \\ 
\textbf{Baseline\cite{Noothout2020}} & 2.32 ± 0.14 & 2.46 ± 0.17 & 2.60 ± 0.11 & 3.18 ± 0.13 \\ \bottomrule
\end{tabularx}
\label{tab:errors}
\end{table}

\begin{figure}
    \centering
    \includegraphics[width=\textwidth, trim={0 0 0 0},clip]{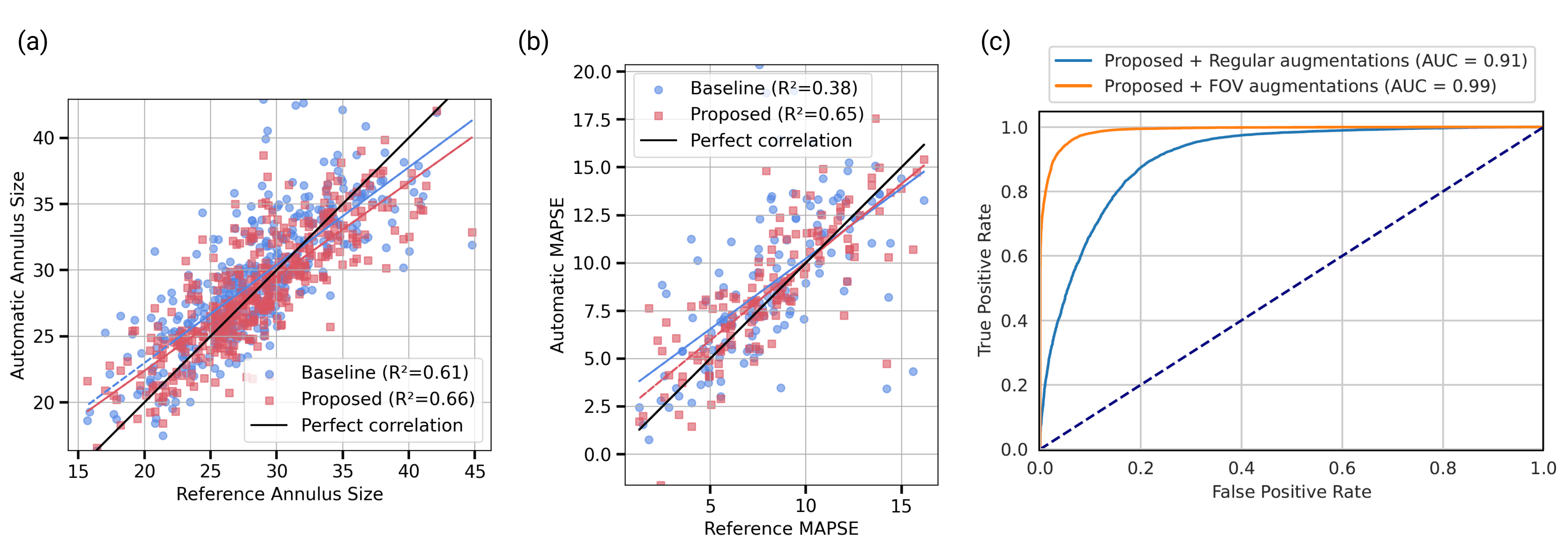}
    \caption{(a, b) Correlation plots of experiments comparing the temporally consistent method with the baseline approach\cite{Noothout2020}.(c) ROC curves for recognition of missing landmarks.}
    \label{fig:corr_roc}
\end{figure}

\begin{figure}
    \centering
    \includegraphics[width=\textwidth, trim={0 0 0 0}, clip]{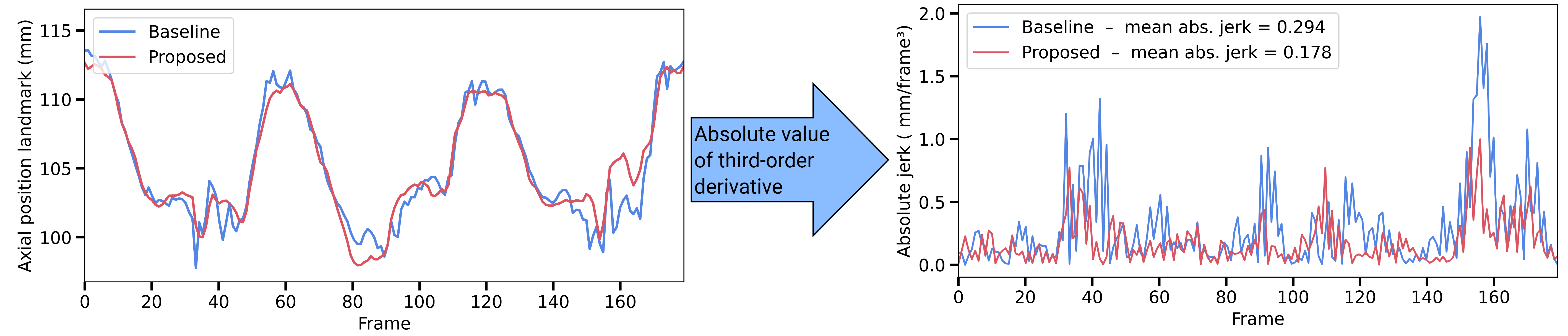} 
    \caption{Left: Axial position of the left annulus landmark plotted over time for a randomly selected test video, comparing the smoothness of the baseline \cite{Noothout2020} and the proposed method. Right: Absolute jerk (third-order derivative) over time, illustrating reduced jerk values for the proposed method. For this visualization, jerk is computed from axial displacement only, while the values reported in Table \ref{tab:errors} are based on the 2D position.}
    \label{fig:jerk}
\end{figure}

\begin{figure}
    \centering
    \includegraphics[width=0.85\textwidth, trim={0 0 0 0},clip]{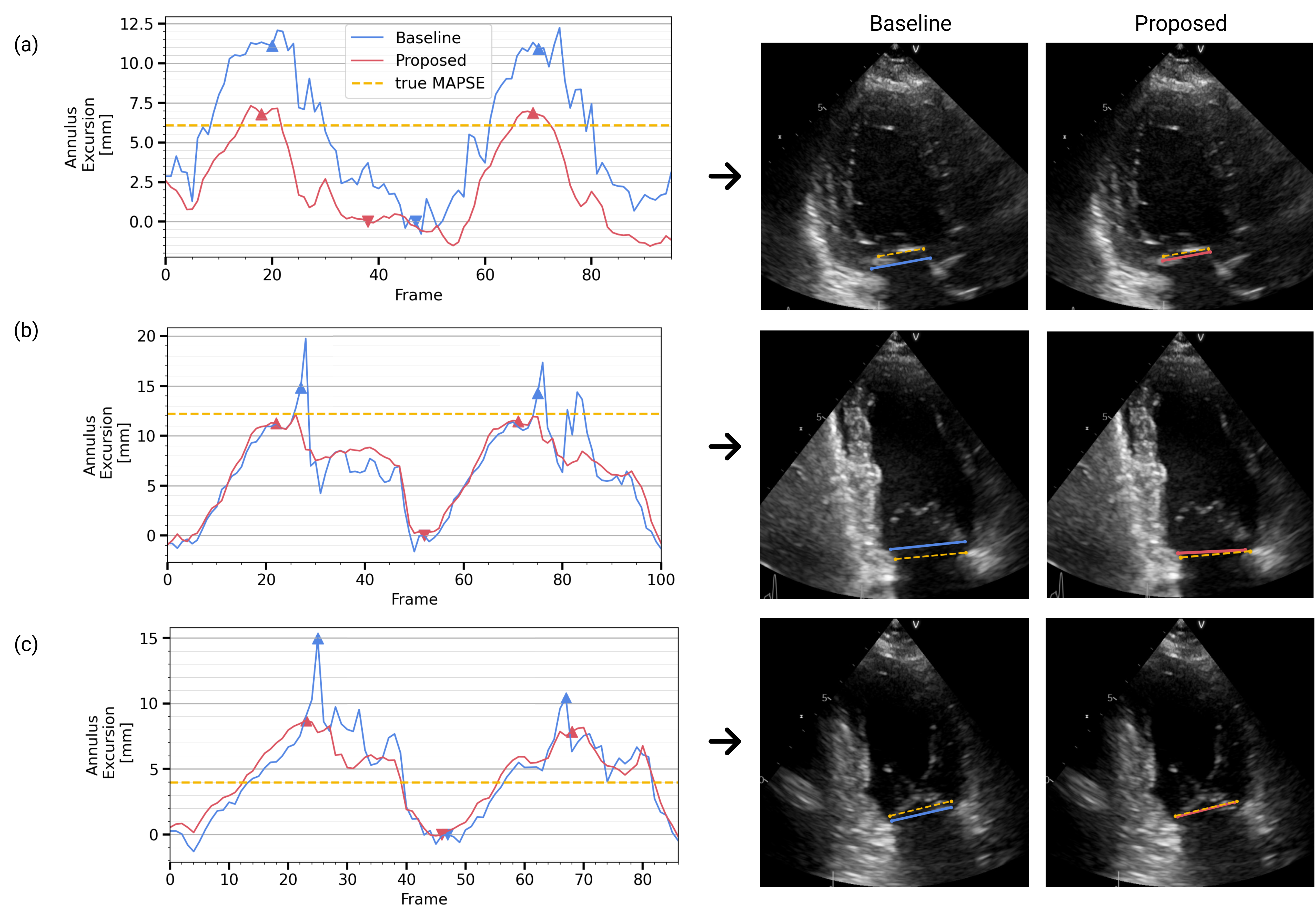}
    \caption{Annulus excursion over time plotted for three test video sequences, comparing the baseline~\cite{Noothout2020} with the proposed method (a–c). The plots display the annulus excursion (in mm) per frame, with the predicted end-diastole (ED, downward triangles) and end-systole (ES, upward triangles) indicated. The maximum annulus excursion (MAPSE) is represented by the dashed line for the reference method, while the predicted MAPSE corresponds to the excursion value at ES (excursion at ED is defined as 0). The accompanying ultrasound frames show the predicted annulus (solid line) and the reference annotation (dashed line) for both methods.}
    \label{fig:mapse_results}
\end{figure}

\subsection{Results}
The proposed method achieved a 2.48 ± 0.07 mm MAE for landmark localization, a 1.81 ± 0.14 mm MAE for MAPSE, and 2.46 ± 0.31 MAE for annulus size on the combined dataset. For comparison, previous work addressing annulus landmark detection reported mean landmark errors of 3.0 and 4.1 mm (for end-diastole and end-systole, respectively) \cite{smistad2022tracking}. To investigate the effect of using the temporal consistency loss term, we compared our method to the (global-only) baseline approach by Noothout et al. \cite{Noothout2020}. Summarized statistics are shown in Table \ref{tab:errors}, which indicate that the proposed method most notably improves the MAPSE measurement and jerk, reducing the mean error from 2.32 to 1.81 mm, and jerk from 3.18 to 0.98 mm/frame$^3$. Correlation plots of annulus size and MAPSE are provided in Figure \ref{fig:corr_roc}(a,b), demonstrating improved correlation between predicted and reference values for the proposed method.
Qualitative results in Figure \ref{fig:jerk} and Figure \ref{fig:mapse_results} demonstrate smoother evolution over time for landmark position and mitral annulus excursion, respectively, while Figure \ref{fig:mapse_results} also shows improved MAPSE calculation by the proposed method. 

\newpage
To test if the proposed augmentation scheme improves the recognition of missing landmarks, we performed an ablation experiment in which we replaced the proposed augmentations with regular cropping and rotations. As our test set only contained videos with valid in-view landmarks, we additionally created 4 randomly augmented versions of each test image, in the same way that we augmented our training set, thereby causing some landmarks to be moved out of view. 
We measured the missing-landmark recognition performance using the area under the ROC curves (ROC-AUC), whereby each landmark was treated as an independent sample. 
The results, summarized in Figure \ref{fig:corr_roc}(c), show that applying the proposed field-of-view augmentations improved the ROC-AUC from 0.91 to 0.99.

% \begin{figure}[t!]
% \floatconts
% {fig:example2}% label for whole figure
% {\caption{Out-of-view recognition comparison of training with a regular augmentation strategy, and the proposed field-of-view (FOV) augmentations. Evaluated on CAMUS dataset with test-time FOV augmentations.}}% caption for whole figure
% {%
% \subfigure{%
% \label{fig:pic1}% label for this sub-figure
% \includegraphics[width=0.43\linewidth]{images/figure_boxplots_oov_recognition.pdf}
% } % space out the images a bit
% \subfigure{%
% \label{fig:pic2}% label for this sub-figure
% \includegraphics[width=0.54\linewidth]{images/figure_roc_oov_recognition.pdf}
% }
% }
% \end{figure}

\section{Discussion}

We presented a method for temporally consistent localization of anatomical landmarks in echocardiography with robustness to missing landmarks outside the field of view. The approach leveraged a fully convolutional neural network that performed both landmark classification and regression \cite{Noothout2020}, thus predicting both patch-wise probabilities and distances. To achieve temporal consistency, we provided three consecutive frames as context to a 2D CNN for landmark detection, and added a self-supervised loss term to enforce consistency between neighboring frames. To improve robustness to missing anatomical landmarks, we introduced realistic field-of-view augmentations. We trained and evaluated this method using two datasets from different sources. 

The neighboring-frame consistency loss contributed to the accuracy of MAPSE calculation in cardiac ultrasounds, and resulted in a smoother evolution of the annulus landmarks over time (Figure \ref{fig:jerk}), better aligning with expected physiological behavior. In our experiments, MAPSE is measured at ED and ES along an averaged normal vector. The time component required for MAPSE calculation may be a factor in the observed improvement with the temporally consistent method over the baseline method\cite{Noothout2020}. The average MAPSE error improvement of about 0.5 mm could improve risk stratification and clinical decision-making for cases close to the clinically relevant cut-off value of 9 mm, which has been linked to increased mortality risk\cite{Romano2017}.

Additionally, we showed that realistic augmentations, which randomly crop out landmark targets in ultrasound images, improve recognition of missing landmarks. 
Although missing landmarks are a common occurrence in clinical data, we should note that the test set for this experiment did not include such data, but, instead, consisted of test-time augmented videos. In future work, we will collect a clinical dataset with missing anatomical landmarks to evaluate its performance in a realistic scenario.
 
As previous work \cite{smistad2022tracking} addressing annulus landmark detection reported higher landmark localization errors, our method may outperform previous work, but direct comparison remains challenging, as different datasets were employed. In the broader context of video landmark detection, further research is needed to investigate whether the proposed method improves the state of the art on other video-based landmark detection tasks.

\section{Conclusion}

We presented a novel method for annulus landmark detection in echocardiography videos, achieving improved localization accuracy, temporal consistency, and robustness to out-of-view landmarks. These led to improved MAPSE measurements, potentially enabling more effective risk stratification of patients with reduced left-ventricular function.

% Our method for temporally consistent localization of anatomical landmarks in echocardiography improves MAPSE calculation and landmark tracking stability. By incorporating temporal consistency and simulating out-of-view landmarks with data augmentation, we address key challenges in echocardiography video landmark detection. 

% Future work will focus on evaluating the method on clinically relevant out-of-distribution data, and evaluating the method on other video landmark detection tasks in echocardiography.

% \section{New or Breakthrough work to be presented}

% We present a method for the localization of anatomical landmarks in 2D echocardiography videos. Novel contributions include the introduction of a self-supervised loss term to learn temporal consistency, and realistic field-of-view augmentations to increase robustness to missing landmarks.

\section{Acknowledgments}
This work was supported by Pie Medical Imaging BV and Esaote SpA.

\bibliography{main} % bibliography data in report.bib
\bibliographystyle{spiebib} % makes bibtex use spiebib.bst

\end{document}